# ConvGRU in Fine-grained Pitching Action Recognition for Action Outcome Prediction

Tianqi Ma, Lin Zhang, Xiumin Diao and Ou Ma

*Abstract*—**Prediction of the action outcome is a new challenge for a robot collaboratively working with humans. With the impressive progress in video action recognition in recent years, fine-grained action recognition from video data turns into a new concern. Fine-grained action recognition detects subtle differences of actions in more specific granularity and is significant in many fields such as human-robot interaction, intelligent traffic management, sports training, health caring. Considering that the different outcomes are closely connected to the subtle differences in actions, fine-grained action recognition is a practical method for action outcome prediction. In this paper, we explore the performance of convolutional gate recurrent unit (ConvGRU) method on a fine-grained action recognition tasks: predicting outcomes of ball-pitching. Based on sequences of RGB images of human actions, the proposed approach achieved the performance of 79.17% accuracy, which exceeds the current state-of-the-art result. We also compared different network implementations and showed the influence of different image sampling methods, different fusion methods and pre-training, etc. Finally, we discussed the advantages and limitations of ConvGRU in such action outcome prediction and fine-grained action recognition tasks.**

***Index Terms***— **fine-grained action recognition, action outcome prediction, deep learning, GRU, ConvGRU**

## I. Introduction

Human action recognition is an active research area in computer vision and human-robot interaction (HRI). Efficiently detecting and understanding human actions can largely boost a robot's performance to collaboratively work with humans in HRI tasks. Such advantages have been demonstrated in various HRI fields, such as sports, healthcare, traffic management [1][2][3][4] and manufacturing industry [5]. Deep learning methods have proved their values of being state-of-the-art solutions in a variety of action recognition tasks [6][7][8][9].

In the early age of the research on action recognition, researchers mainly aim at coarse-grained action recognition, e.g. "riding a horse" vs "playing a piano" [10]. The background context often provides extra information to help the recognition of performance [11]. In some extreme conditions in application, however, the actions are more specific. An intelligent pitching-catching training system, for instance, values the ability of detecting the goal of the pitching ball from the pitching action of a participant more than distinguishing "pitching" from "jogging" or "boxing". Therefore, a robot can benefit more from the ability of detecting and recognizing human actions in more specific granularity, which is named fine-grained action recognition.

Prediction of the action outcome is a practical application of fine-grained action recognition. It aims at predicting the different outcome of one action. We can distinguish different actions to predict different outcomes. The reason is that even for one kind of action, two totally same actions in the same circumstance must result in the same outcome. If the outcome is different, there must be some differences (usually subtle) in the actions. The mapping from actions to the corresponding outcome, however, is complex but definite. For example, a person pitches a ball and the ball will land at different points with different pitching velocities, positions and postures. Thus, fine-grained action recognition helps us detect the subtle differences of the actions and also the different outcomes behind them.

A challenging problem of fine-grained human action recognition is that few differences in input may lead to many different outcomes [12][13][14]. For example, how to infer a person's intention of grasping an object from the table? Obviously, the main actions are same but the slight variations will cause the very different outcomes (e.g. your palm could be down if you take an orange on the table, but different when the object changes to a cup. The other parts of the action "grasping" are almost the same). To extract the subtle differences from actions, researchers used precision sensors such as wearable sensors [15] to capture action details and devices to catch participants' gazes [16]. They even used biological signals [17][18] to precisely represent the actions. At the same time, multi-sensors [1] also improve the ability to extract features.

However, dependence of special and expensive sensors limits the applications and may also discomfort the participants. A person will unlikely conduct action naturally while wearing uncomfortable or distractive sensors. With the development of convolutional neural networks (CNN), action recognition based on one camera and videos becomes a significant method to this problem. To extract spatial and temporal features in fine-grained action videos, researchers use two-stream [13][19][20] or multi-stream [21][22] methods, 3D-CNN [23][24] and LSTM (Long short-term memory) [25][26]. The CNN method also provides an end-to-end model.

In this paper, we explore the performance of convolutional gate recurrent unit (ConvGRU) [27] in fine-grained action recognition tasks, which has not been well studied to the best of our knowledge. To fully explore ConvGRU, we built the network based on AlexNet [28] and attempted different network implementations. We carried out experiments on our previous open source pitching dataset [13], which is a fine-grained dataset based on human pitching a ball and the pitching actions are labeled by different pitching goal blocks. We achieved the maximum of 79.17% accuracy in prediction of landing blocks of human pitching actions. Experiment results also demonstrated that:

1) The recognition accuracy can be improved by: a) proper sampling interval, b) longer sampling length, c) early fusion and d) bi-directional structure.
2) Pre-trained models with fine-tuning parameters do not always assist the fine-grained action recognition.
3) ConvGRU is robust against spatial-temporal errors, where it is more robust to temporal implementation errors than spatial errors.

The rest of this paper is organized as follows. In Section II, the related research about action recognition is described. Section III introduces the details about our network architecture. Then in Section IV, the dataset and experiment implementation are discussed. The results and discussions are in Section V. Finally, the paper is concluded with a discussion in Section VI.

## II. Related Work

**Methods for video action recognition.** The differences between action recognition from videos and still images are temporal features, which exist only in the video data. To extract temporal features and properly fuse them with spatial features, the applicable deep learning methods can be divided into the following three categories: 1) Based on 2D-CNN [29][30] with two-stream [13][19][20][31][32] or multi-stream [21][22] inputs. Raw RGB images and optical flow [33] are often selected as the two-stream input and represent spatial and temporal features respectively. Among the multi-stream methods, the other inputs vary based on different targets to be recognized. For example, in [22], the authors separately divide RGB image flow and optical flow in three parts: full, human and human operation then build a six-stream network. 2) Based on 3D-CNN [23][24][34][35]. This category uses 3D kernels instead of 2D kernels in CNN and the extra dimension operates on time dimension to extract temporal features. 3) Based on LSTM [25][26][36][37]. The methods based on LSTM can be roughly divided into two classes. The first one is to first extract image features, and then use LSTM for recognition. For example, Donahue et al. [38] proposed long-term recurrent convolutional networks (LRCN) which directly encode each image in video and then recognize the action with LSTM. The second class of LSTM based methods is to blur the boundary between feature extraction and temporal feature processing. The SCNN model [39] can directly encode the spatial and temporal dimension of image together with convolutional neural networks (CNN).

**Datasets for fine-grained action recognition.** Datasets for coarse-grained action recognition have been widely explored, such as UCF-101 [40], HMDB-51 [41], Kinetics [42] and ActivityNet [43]. Previous researches have reached good recognition accuracy on these datasets. There are also significant attempts on fine-grained action datasets [11][13][14][44][45][46][47][48][49]. However, to our best knowledge, none of them has been widely accepted as a baseline similar to UCF-101 and HMDB-51 in coarse-grained action dataset. Different from the various types of actions in coarse-grained datasets, fine-grained datasets usually aim at one scene or one action. For example, FineGym [11] includes 4 events in gymnastics and provides multi-semantic annotations. MPII-Cooking 2 [47] consists of more than 200 videos of cooking activities in the kitchen. In this paper, we conduct our experiments on the dataset provided in [13], which aims at human pitching a ball to different goal blocks.

## III. ConvGRU and Network Architecture

In this section, we will first briefly describe ConvGRU. Then, we will show our end-to-end network architecture for fine-grained recognition of pitching action.

### A. ConvGRU

Gated Recurrent Unit (GRU) [50] is provided first for machine translation. GRU includes gates in one unit as LSTM does, but the structure is simpler. Compared to LSTM, GRU reduces a gate structure but performs similarly as LSTM, thus GRU is computational cheaper. ConvGRU (as shown in Figure 1) combines CNN and GRU, thereby maintaining the spatial structure of the input and more conducive to extracting spatial-temporal features in the video. The notion comes from SCNN [39]. The basic math principle of ConvGRU is [27]: with an action image sequences $X$ with $T$ frames $X = \{x_1, \dots x_T\}$, where $x_i \in \mathbb{R}^{N\times \mathrm{N}}$, it performs the forward propagation for all $t = 1, \dots, T$:

$$z_t = \sigma(w_{zx} * x_t + w_{z\mathrm{h}} * h_{t-1} + b_z) \quad (1)$$
$$r_t = \sigma(w_{rx} * x_t + w_{rh} * h_{t-1} + b_r) \quad (2)$$
$$o_t = tanh(w_{ox} * x_t + w_{oh} * (r_t \odot h_{t-1}) + b_o) \quad (3)$$
$$h_t = z_t \odot x_t + (1 - z_t) \odot o_t \quad (4)$$

where operators $*$ and $\odot$ respectively represent convolution operation and Hadamard product; $tanh(\cdot)$ and $\sigma(\cdot)$ respectively represent hyperbolic tangent and activation function (Sigmoid function in general). As Fig.2 illustrates, $w_x \in \mathbb{R}^{d\times d}$ , $w_h \in \mathbb{R}^{1\times 1}$ and $b \in \mathbb{R}^{n\times n}$ represent the corresponding weights and biases (the relationship between the dimensions before and after convolutional is shown in Table 1 and other operations in the formulas do not change dimensions). For each time index $t = 1, \dots, T$ , $h_t \in \mathbb{R}^{n\times n}$ means hidden state, which can be regard as both output and historical information flowing in the network. $z_t \in \mathbb{R}^{n\times n}$, $r_t \in \mathbb{R}^{n\times n}$ and $o_t \in \mathbb{R}^{n\times n}$ represent different gates in GRU. We assume $h_0 = \mathbf{0}$ as the initial condition.

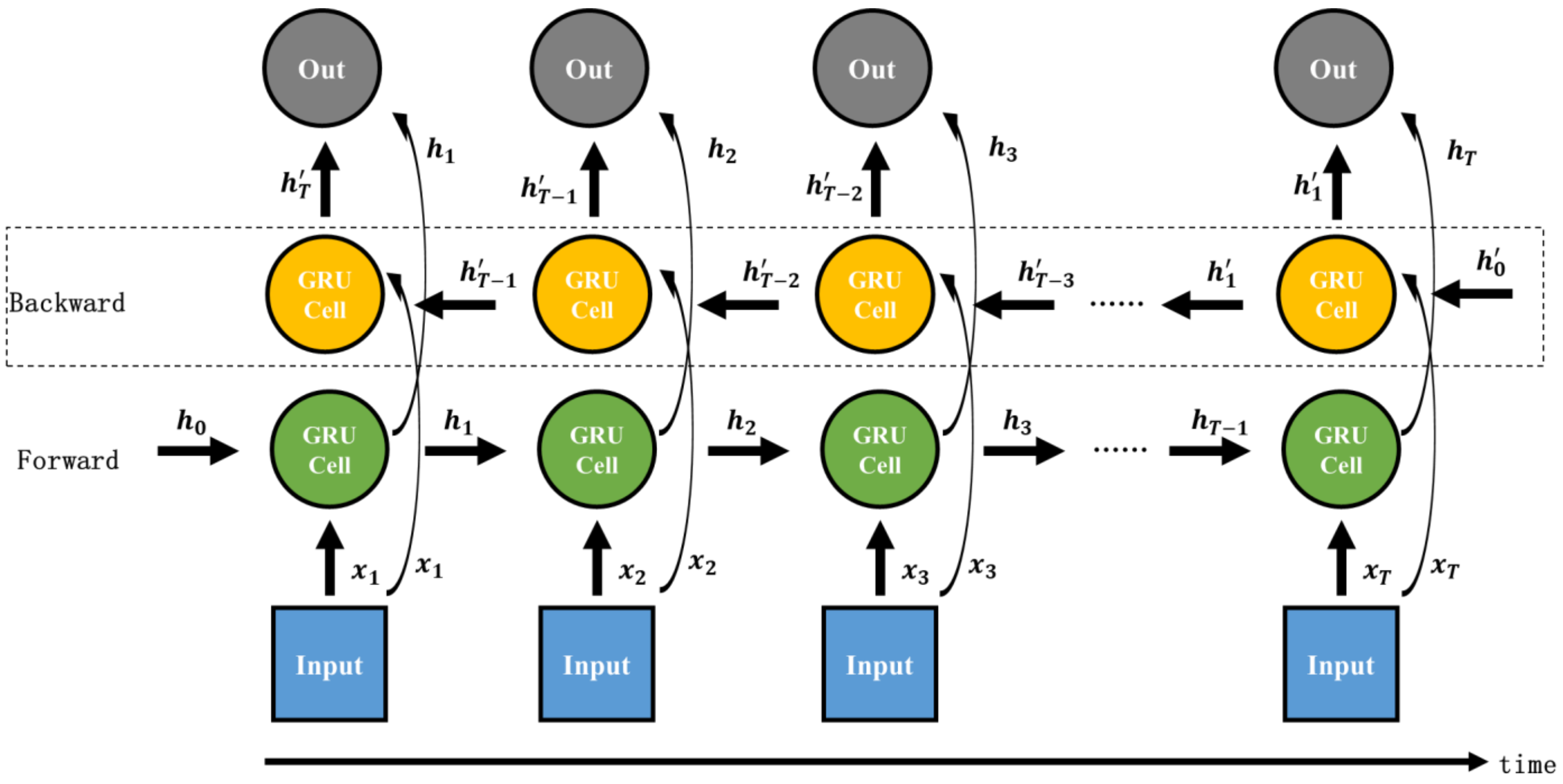

Fig.1. General ConvGRU layer. Backward ConvGRU in the dotted box only exists if bi-directional ConvGRU is applied.

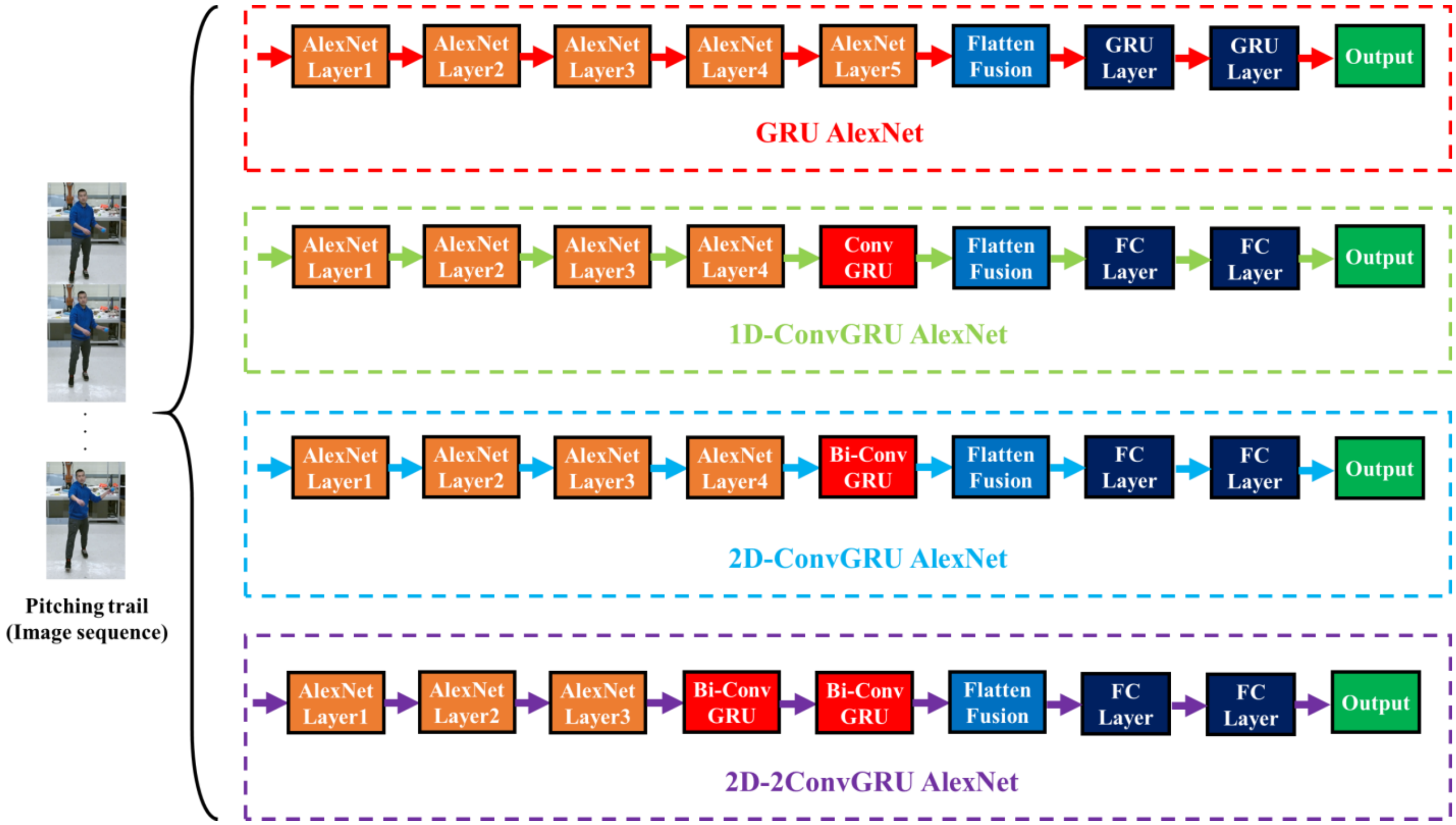

Fig.2. The structure of proposed models

Similar to BI-LSTM [51], bi-directional GRU adds an extra GRU structure in parallel on the basis of original GRU, but in the opposite direction. Two GRU are not affected by each other during the propagating, but the final output of each frame is jointly determined by the outputs of the two GRUs. With the addition of a backward GRU, the temporal extracting characteristic of bi-directional GRU also changes. This is because each output is jointly determined by a forward GRU and backward GRU. A GRU is able to synthesize the current and historical input. Therefore, besides integrating historical information in the forward GRU, we can obtain "historical" information in the backward GRU, which is in fact the future information. In other words, the bi-directional GRU is a combination of current, historical and future information. Similarly, we have backward propagation formulas (we use the hat mark ^ to represent variables in the backward GRU to distinguish them from these in the forward GRU):

$$\hat{z}_t = \sigma(\hat{w}_{zx} * x_t + \hat{w}_{zh} * \hat{h}_{t+1} + \hat{b}_z) \quad (5)$$

$$\hat{r}_t = \sigma(\hat{w}_{rx} * x_t + \hat{w}_{rh} * \hat{h}_{t+1} + \hat{b}_r) \quad (6)$$

$$\hat{o}_t = tanh(\hat{w}_{ox} * x_t + \hat{w}_{oh} * (\hat{r}_t \odot \hat{h}_{t+1}) + \hat{b}_o) \quad (7)$$

$$\hat{h}_t = \hat{z}_t \odot x_t + (1 - \hat{z}_t) \odot \hat{o}_t \quad (8)$$

The backward propagation starts from $t = T$ to $t = 1$ and we assume $h_{T+1} = \mathbf{0}$ being the initial condition. Equations (1) through (8) are the bi-directional ConvGRU propagation formulas.

For each time index $t$, we will get the forward (backward) GRU outputs $h_t$ ($\hat{h}_t$), then where to fuse the output remains a question. In our experiments, we consider 4 fusion methods: 1) reshape the last output $h_T$ as a 1D vector (last_flat); 2) reshape the mean output of all inputs of sequences $h_1 \dots h_T$ as a 1D vector (mean_flat); 3) apply average pooling to the last output $h_T$ to acquire a 1D vector (last_avg); and 4) reshape each $h_t, t = 1, \dots, T$ as a 1D vector separately and fuse them after full connection layer (flat). If the bi-directional GRU is used, we will replace $h_T$ with $(\hat{h}_1 + h_T)/2$ in the fusion schemes 1) and 3) and replace $h_t$ with $(\hat{h}_t + h_t)/2$ in the fusion schemes 2) and 4).

### B. *Modified AlexNet based on ConvGRU*

The original architecture of AlexNet consists of 5 convolution layers and 2 fully connection layers. It also applies ReLU (rectified linear unit), dropout and local response normalization (LRN) to improve the effect. We construct 3 improved networks with ConvGRU and 1 improved network with GRU in full connection layer as shown in Table 1 and Fig.2.

TABLE I
NETWORK ARCHITECTURE

| Layer Number | 1D-ConvGRU AlexNet | 2D-ConvGRU AlexNet | 2D-2 ConvGRU AlexNet | GRU AlexNet |
|---|---|---|---|---|
| 1 | 11×11 conv, 96 features | | | |
| 2 | 3×3 max pooling, stride 2 | | | |
| 3 | 5×5 conv, 256 features | | | |
| 4 | 3×3 max pooling, stride 2 | | | |
| 5 | 3×3 conv, 384 features | | | |
| 6 | 3×3 conv, 384 features | 3×3 conv, 384 features | 3×3 bi-ConvGRU, 384 features | 3×3 conv, 384 features |
| 7 | 3×3 ConvGRU, 256 features | 3×3 bi-ConvGRU, 256 features | 3×3 bi-ConvGRU, 256 features | 3×3 conv, 256 features |
| 8 | fusion and flatten | | | |
| 9 | full connection, 4096 hidden units | | | GRU, 1024 hidden units |
| 10 | full connection, 4096 hidden units | | | GRU, 1024 hidden units |
| 11 | output | | | |

If input is $N \times N$ and convolution kernel is $d \times d$, the dimension of the output $n \times n$ is dependent on the following two factors: 1) stride and 2) padding mode (SAME or VALID). Assuming that the stride is $s$ for both the $x$ and $y$ directions, then

$$n = \begin{cases} \text{ceil}\left(\frac{N}{s}\right), \text{if SAME} \\ \text{ceil}\left(\frac{N-d+1}{s}\right), \text{if VALID} \end{cases}$$

## IV. DATASET AND EXPERIMENT

In this study, we used the dataset proposed in [13], which contains sequential images of human participants performing ball pitching task as shown in Fig.3. The dataset collected data from 6 participants and the target was divided into 9 blocks as different sub-targets. Each participant pitched a ball into each target block 10 times. Therefore, we have 6×9×10=540 pitching trials in total. The researchers used Kinect V2 sensor (Microsoft Corporation, Redmond, WA, USA) to collect data and the frame rate was set to default (30 frames a second). Each trial lasted 3 seconds and thus resulting a total of 90 frames. RGB images, depth images and anatomical joint positions were all collected.

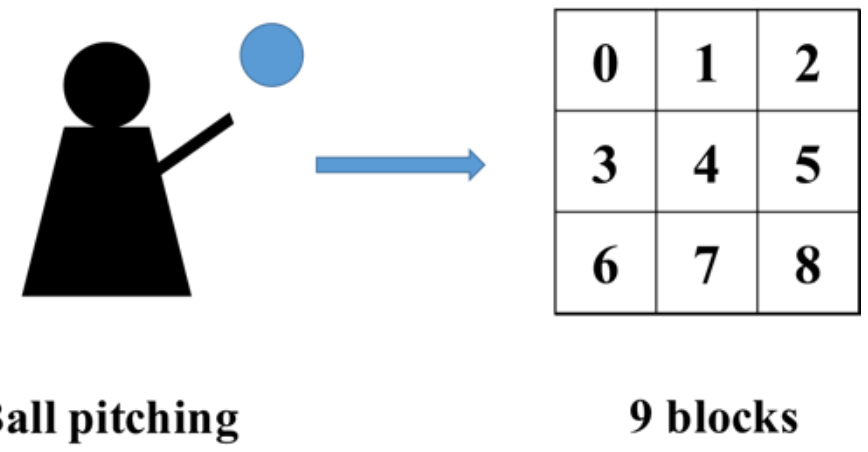

Fig.3. Experiment scenario in [13]

### A. *Starting frame detection*

Before training, we need to first determine the effective action interval. This is because the original data in one trial contains useless frames before and after (as shown in Fig.4). Since multiple participants were involved and they had different motion patterns, the starting frame of each trial varied from participant to participant. It is necessary to select the right starting frame for each trial instead of just simply picking a fixed starting frame for all trials. In order to avoid local minimum, we use RGB images instead of joint coordinates in [13] as discriminant criteria to detect starting frame in a pitching trial. We assume that the pitching does not start in the first 10 frames, which is reasonable on the available dataset. For a trial with joint positions $\theta_t$ and normalized, central-cropped RGB images $x_t$ from $t = 1$ to $T$, we have

1) Calculate $\Delta\theta = [\Delta\theta_{11}, \dots, \Delta\theta_T]$ , where $\Delta\theta_t = \|\theta_t - \theta_{10}\|_2, t = 11, \dots, T$ and $\Delta x = [\Delta x_{11}, \dots, \Delta x_T]$ , where $\Delta x_t = \|x_t - x_{10}\|_F, t = 11, \dots, T$.
2) Set thresholds $\delta_1 = 6$ and $\delta_2 = 0.16$. Find minimum $i$ which satisfies a) $\Delta x_t \geq \delta_1, t = i, \dots, i+7$, b) $\Delta x_{t+1} \geq \Delta x_t, t = i, \dots, i+6$ and c) $\Delta\theta_i \geq \delta_2$.
3) Set frame $i$ as the starting frame.

Experiment results show that all trials have a frame which satisfies a), b) and c) in 2).

We use a 40-frame interval from the starting frame for the training and testing processes.

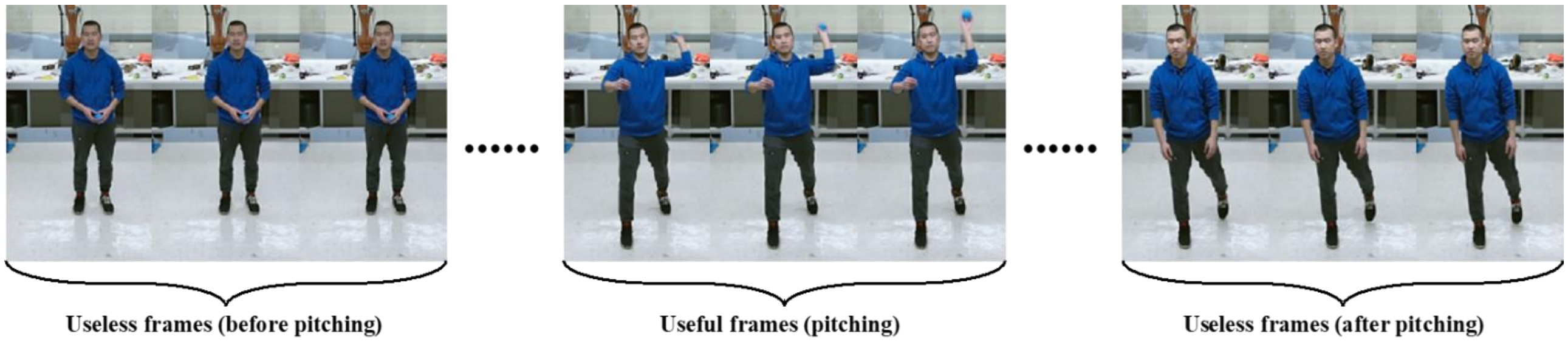

Fig.4. Useless frames before and after action

## B. *Experiment Implementation*

**Sampling.** After determining the start frame, each trial will have 40 frames, which is still too much for training, and thus we sample images as the input. Considering that ConvGRU learns features from time-sequence images, we sample images at a fixed or nearly-fixed time interval. We will explore the sampling method for best performance. The method is chosen from Table 2.

**Network inputs.** For each image for training, data augmentation is applied. We first crop with a size $240 \times 240$ from the central part. Then we randomly crop with a size $235 \times 235$ and resize to $224 \times 224$. Next, we randomly change the brightness, contrast, saturation and hue of the image. Finally, we regularize the pixels.

**Hyper-Parameter.** We set the max training epoch as 300, the learning rate as 0.0001 and the batch-size as 32. Dropout rate is 0.8. All activation functions are leaky ReLU and the parameter α is 0.1. The parameters of data augmentation of brightness, contrast, saturation and hue are 0.5, 0.5, 0.5, and 0.1 respectively. We choose Adam optimizer [52] with L2-norm regularization with a weight 0.0005 to train our networks.

**Pre-trained model.** We compare the models with pre-trained parameters on ImageNet and without pre-trained parameters.

TABLE II
SAMPLING METHODS

| | Sampling for training ({first frame}, step, length) | Sampling for testing (mean of results of all sampling) |
|---|---|---|
| 1 | {0, 1}, 2, 20 | {0, 1}, 2, 20 |
| 2 | {0, 1, 2, 3}, 4, 10 | {0, 1, 2, 3}, 4, 10 |
| 3 | {0, 1, 2, 3, 4}, 5, 8 | {0, 1, 2, 3, 4}, 5, 8 |
| 4 | {0, 1, 2, 3, 4, 5, 6, 7}, 8, 5 | {0, 1, 2, 3, 4, 5, 6, 7}, 8, 5 |
| 5* | {0, 1, 2, 3}, {3, 4, 5}, 10 | {0, 1, 2, 3}, 4, 10 |
| 6 | {0, 1, 2…, 29},1,10 | {0,10,20,30}, 1, 10 |
| 7 | {0, 1, 2…, 19},2,10 | {0,1,20,21}, 2, 10 |

**Example (sampling method 2):** Take the starting frame as frame 0. While training, for a trial, we randomly choose frame from {0, 1, 2, 3} as the first frame, and sample a 10-images-sequence with step 4 to train our network. While testing, we will divide all 40 frames into 4 10-images parts, with frame 0, 1, 2, 3 as the first frame and step 4. Test results comes from the mean of these 4 sequences.
* Sampling is limited in 40 frames

**Test**. We choose 8 out of 10 trials of each participant as the training set, and the rest 2 as the testing set. Therefore, the size of the training set and testing set are 432 and 108, respectively. Different from images for training, we crop with a size $240 \times 240$ from the central part and resize to $224 \times 224$. Finally, we regularize the pixels. Other data augmentation methods are not applied here.

# V. RESULTS

## A. *Overall training results*

We conducted experiments using Pytorch on a computer running Windows 10 with an AMD Ryzen 9 4900H, 16 GB RAM, and an NVIDIA 2060 GPU card, 6GB VRAM. Each model is trained twice and the mean performance is recorded. The overall results are shown in Table 3. Our best result shows an apparent improvement compared to the baseline of two-stream (**79.17%** vs 71.30%) in [13] and the original GRU AlexNet (**79.17%** vs 65.74%).

TABLE III
OVERALL TRAINING RESULTS

| Model | Accuracy (%) | Mean training time for each epoch (s) |
|---|---|---|
| Baseline (spatial only in [13]) | 63.89 | / |
| Baseline (temporal only in [13]) | 74.07 | / |
| Baseline (two-stream fusion in [13]) | 71.30 | / |
| Baseline (GRU AlexNet) Sampling 1+ flat | 65.74 | 34 |
| 1D-ConvGRU Sampling 1+last_flat | 76.85 | 33 |
| 1D-ConvGRU Sampling 2+last_flat | 75.93 | 19 |
| 1D -ConvGRU Sampling 3+last_flat | 73.15 | 16 |
| 1D -ConvGRU Sampling 4+last_flat | 73.15 | 12 |
| 1D -ConvGRU Sampling 5+last_flat | 74.54 | 19 |
| 2D-ConvGRU | **79.17** | 46 |
| Sampling 1+last_flat | | |
| 2D -ConvGRU<br>Sampling 1+last_flat+pre-train | 69.44 | 46 |
| 2D-ConvGRU<br>Sampling 2+last_flat | 77.31 | 20 |
| 2D-ConvGRU<br>Sampling 2+last_flat+pre-train | 68.52 | 20 |
| 2D-ConvGRU<br>Sampling 3+last_flat | 75.00 | 19 |
| 2D-ConvGRU<br>Sampling 3+last_flat+pre-train | 66.20 | 19 |
| 2D-ConvGRU<br>Sampling 4+last_flat | 73.15 | 13 |
| 2D-ConvGRU<br>Sampling 4+last_flat+pre-train | 70.37 | 13 |
| 2D-ConvGRU<br>Sampling 5+last_flat | 77.78 | 21 |
| 2D-ConvGRU<br>Sampling 5+last_flat+pre-train | 67.59 | 21 |
| 2D-ConvGRU<br>Sampling 1+mean_flat | 68.98 | 47 |
| 2D-ConvGRU<br>Sampling 1+last_avg | 33.33 | 43 |
| 2D-ConvGRU<br>Sampling 1+flat | 74.07 | 53 |
| 2D-ConvGRU<br>Sampling 2+mean_flat | 77.78 | 21 |
| 2D-ConvGRU<br>Sampling 2+last_avg | 47.22 | 21 |
| 2D-ConvGRU<br>Sampling 2+flat | 70.83 | 22 |
| 2D-ConvGRU<br>Sampling 6+last_flat | 65.74 | 21 |
| 2D-ConvGRU<br>Sampling 7+last_flat | 70.37 | 21 |
| 2D-2ConvGRU<br>Sampling 5+last_flat | 75.93 | 28 |

## B. *Performance of GRU and ConvGRU*

The performance of GRU and ConvGRU is shown in Fig.5. First, both GRU (65.74%) and ConvGRU (maximum 79.17%) improve the performance of the spatial-only network (63.89%), which means that GRU structure in the network effectively extract the temporal features. Second, the Bi-directional ConvGRU structure is slightly better than forward-only ConvGRU, however, two bi-directional ConvGRU structures do not show more improvement on the dataset. Therefore, one bi-directional ConvGRU structure suits best for this task.

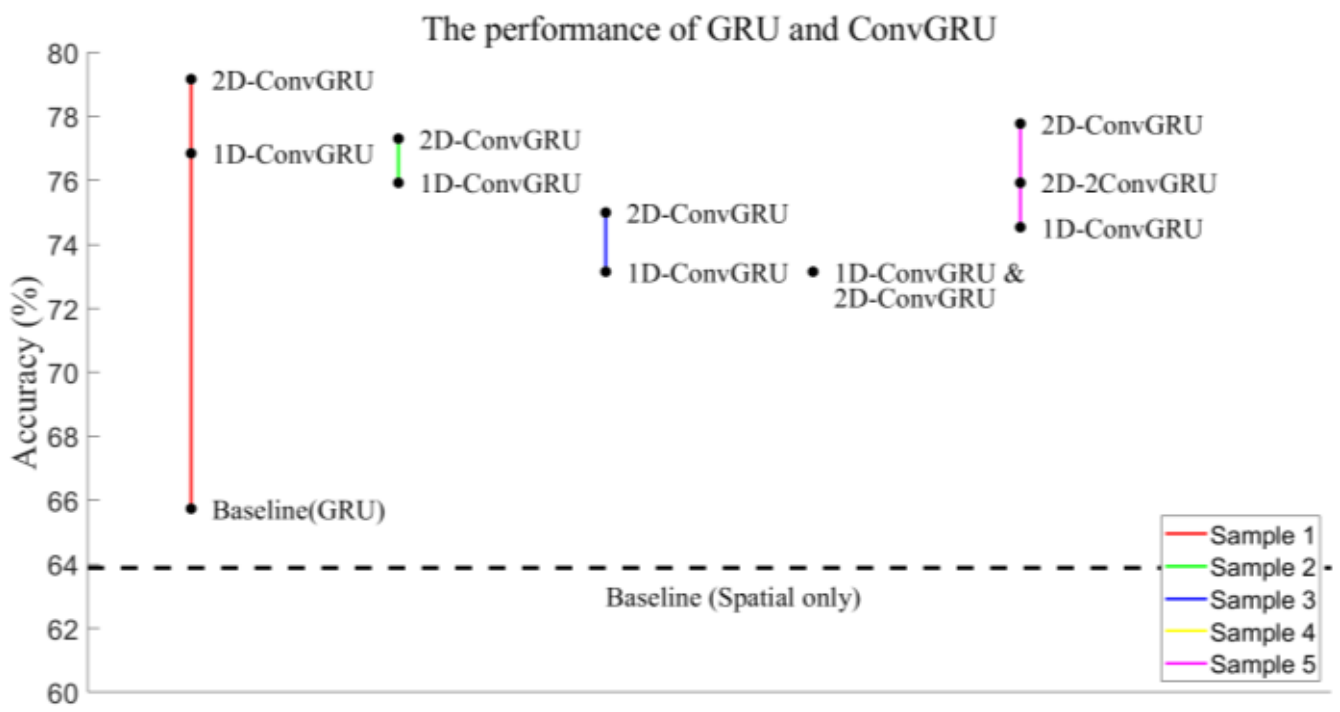

Fig.5. Accuracy of models with different GRU or ConvGRU structures

## C. *Sampling methods*

Figure 6 illustrates the influence of sampling methods. Two main results are shown based on our experiments. First, for ConvGRU, sampling methods that divide the full action into several equal parts and select frames from each part with a fixed or nearly fixed time interval is more feasible. For example, sample method 2, 6 and 7 all sample 10 frames from all 40 frames with different time interval, but sampling method 2 is the best one. We infer the reason is that each sampling in sampling method 2 represents a set of images of full action and the network only needs to learn one pattern of the action. However, sampling methods 6 and 7 sample images with multi-pattern (e.g. In method 7, frame {0, 2, 4…, 18} and {20, 22, 24…, 38} are both legal sampling, however, they represent the former and latter half of the action respectively), and thus is more difficult to learn. Sampling method 6 has more patterns because of shorter time intervals and the accuracy is lower, which supports our inference.

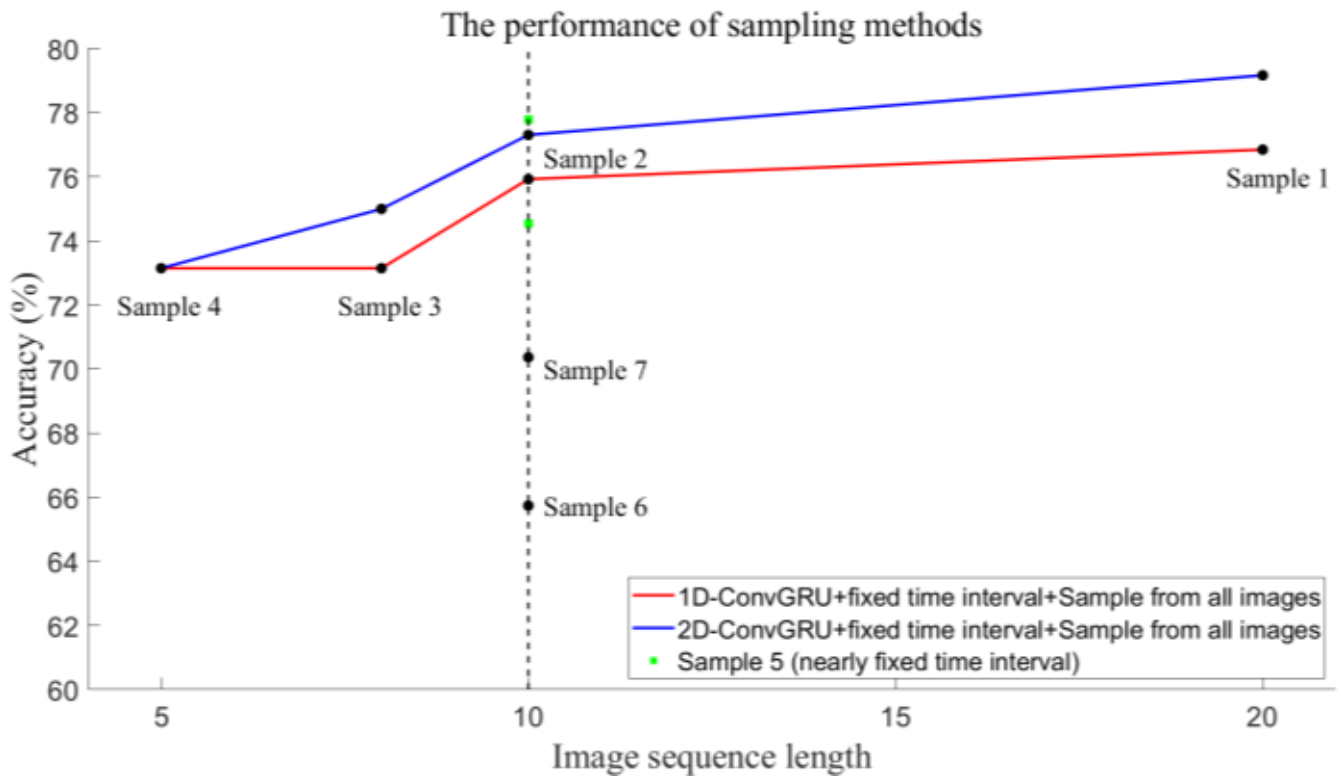

Fig.6. Accuracy of models with different sampling methods

The second conclusion is that longer sampling length leads to higher accuracy in general. We infer that if all 40 frames are used, the performance will be better. However, we also find out that there is only slight improvement when 10-frames changes to 20-frames, but the change requires much more memory and computation. Thus, we do not use the full sequences for input. The third result is that the nearly-fixed sampling has the same or slightly better results than the fixed sampling. That is because the nearly-fixed sampling brings more robustness of time dimension, which we will show in details in the Section F.

In summary, sampling method 1 has the best performance, while sampling method 5 is a computationally low-cost alternative.

### D. Fusion methods

The four fusion methods mentioned in Section III-A can be divided into two kinds: 1) Early fusion (last_flat, last_avg, mean_flat): features of different images are fused before full-connection layers; 2) Late fusion (flat): features are fused after full-connection layers. The performance with different fusion methods is shown in Fig.7. We can find out that last_flat is the best of early fusion methods and outperforms the late fusion method. In early fusion methods, the mean operation may decrease the accuracy.

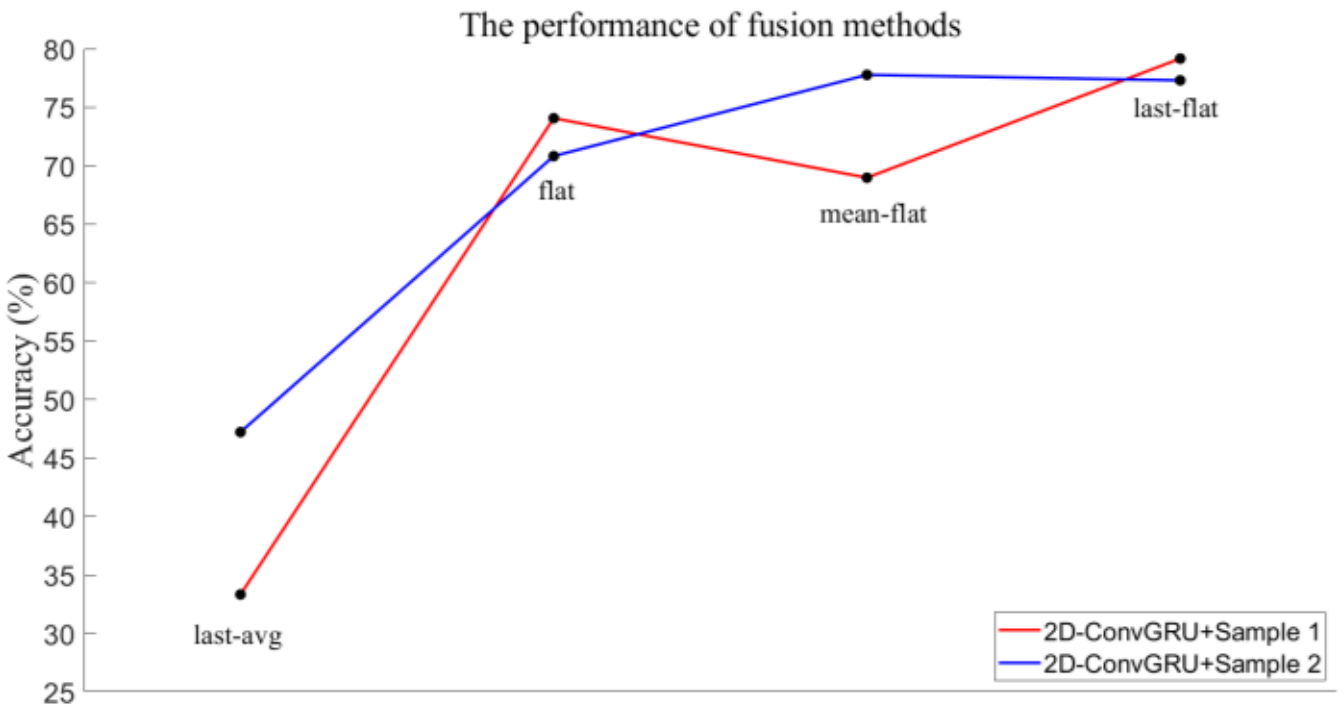

Fig.7. Accuracy of models with different fusion methods

### E. Pre-trained models

Pre-trained models are widely applied in the related researches. However, in fine-grained action recognition, pre-trained models are not always playing a positive role [11]. In our experiments, we also explore the pre-trained models on ImageNet. The results are shown in Fig.8 and our results illustrate that almost all pre-trained models negatively influence the performance. A potential explanation is that the parameters of the model which pre-trained on coarse-grained dataset focus more on background or other parts, and these parameters are not feasible for fine-grained action recognition.

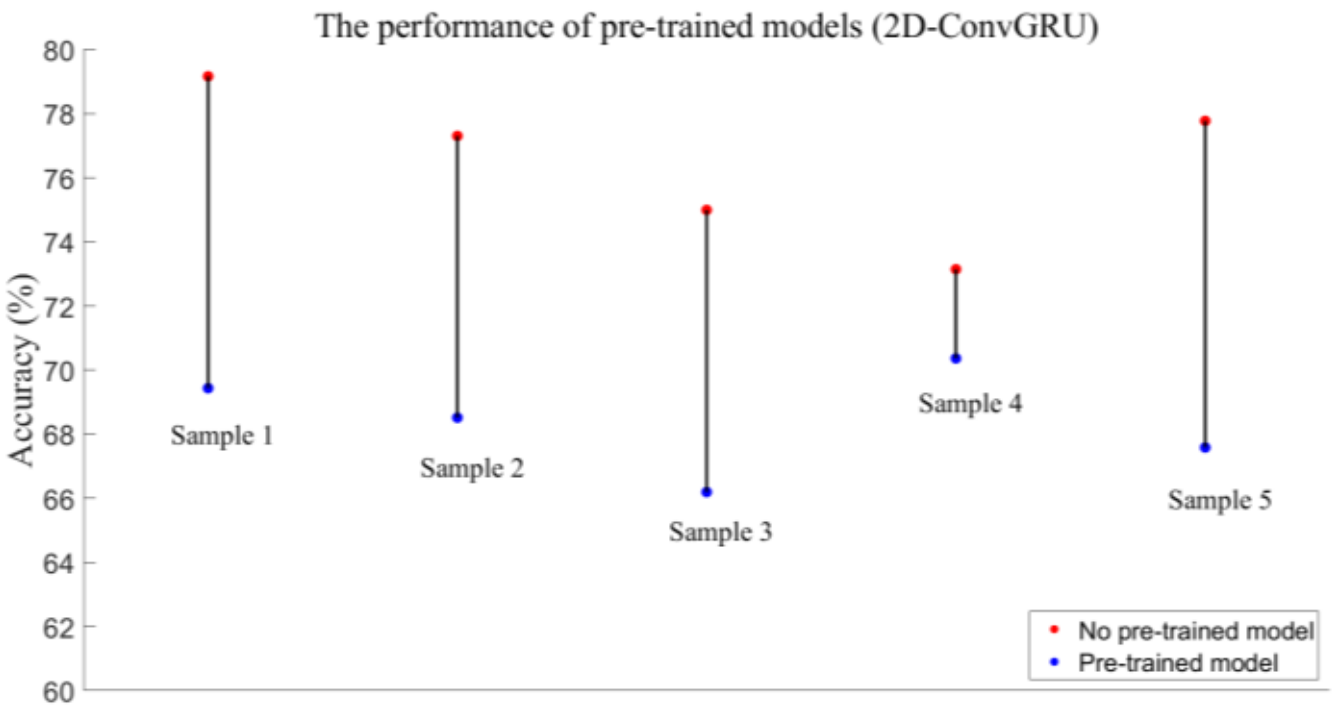

Fig.8. Accuracy of models with pre-trained models

### F. Visualization and Robustness of ConvGRU

We would like to make sure the important factor in the images. In this section, we visualize the output feature of conv_1 layer in the model 2D-ConvGRU+Sample 5+last_flat and the results are shown in Fig.9. The feature maps show that the right half body that participants in the pitching action mainly activates the network (white part in the feature maps). This means that our network not only focuses on the pitching arm/hand, but also all the body parts that help the pitch action, which improves the performance of the network. However, at the same time, the factors of background are not eliminated clearly, because there are still some factors of the background activating the network.

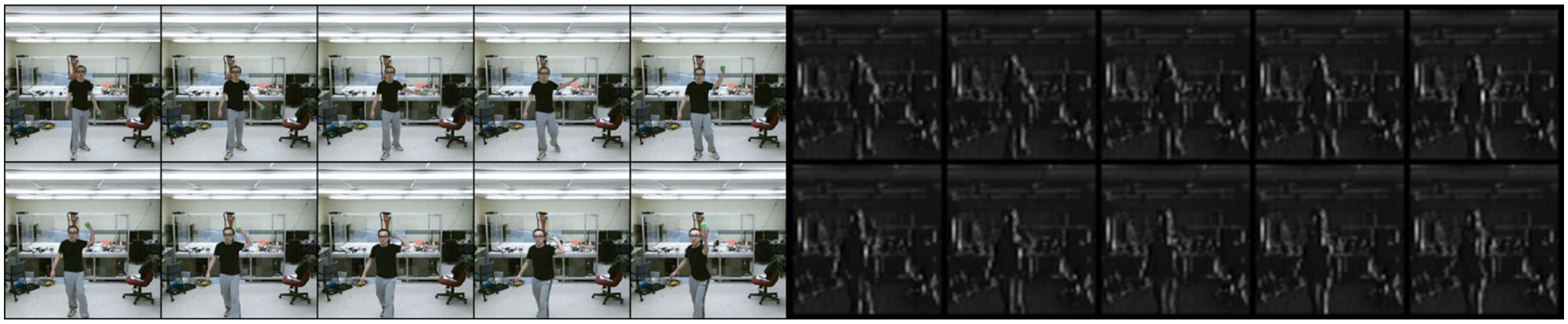

Fig.9. The original images and feature maps of Conv_1 in model 2d-ConvGRU+Sample 5+last_flat

Considering that errors exist while collecting data in real life applications, we will test the robustness of our ConvGRU models. There are 3 situations that we are interested in:

1) Errors in frame rate: our data is 30 frames per second (fps) but in fact, the true frame rate may fluctuate.

2) Missing frames: some images of collected data may be missing.

3) Errors in the position of the participants: participants stood at a nearly-fixed point while collecting the data for the dataset, so the participants in the image are expected to be at the center of the images.

We explore the robustness on model 2D-ConvGRU + Sample 1 + last_flat, 2D-ConvGRU + Sample 2 + last_flat and 2D-ConvGRU + Sample 5 + last_flat. Each test is repeated three times and the mean accuracy is recorded.

Regarding the error in the frame rate, we choose a sequence with more than or less than 40 images (41 to 45) from the starting frame to simulate a frame rate error about 10% downward (26.67 fps to 30 fps) and randomly delete the extra images to build a new 40-images sequence. The results are shown in Fig.10. All the three models have robustness to the rate error and the accuracy does not decrease rapidly. One interesting result is that, compared to model with sampling method 2, the accuracy of model with sampling method 5 even increases with the slight change of frame rate. This is possibly because the latter model is trained with nearly-fixed time intervals and is more feasible to the slight change of frame rate.

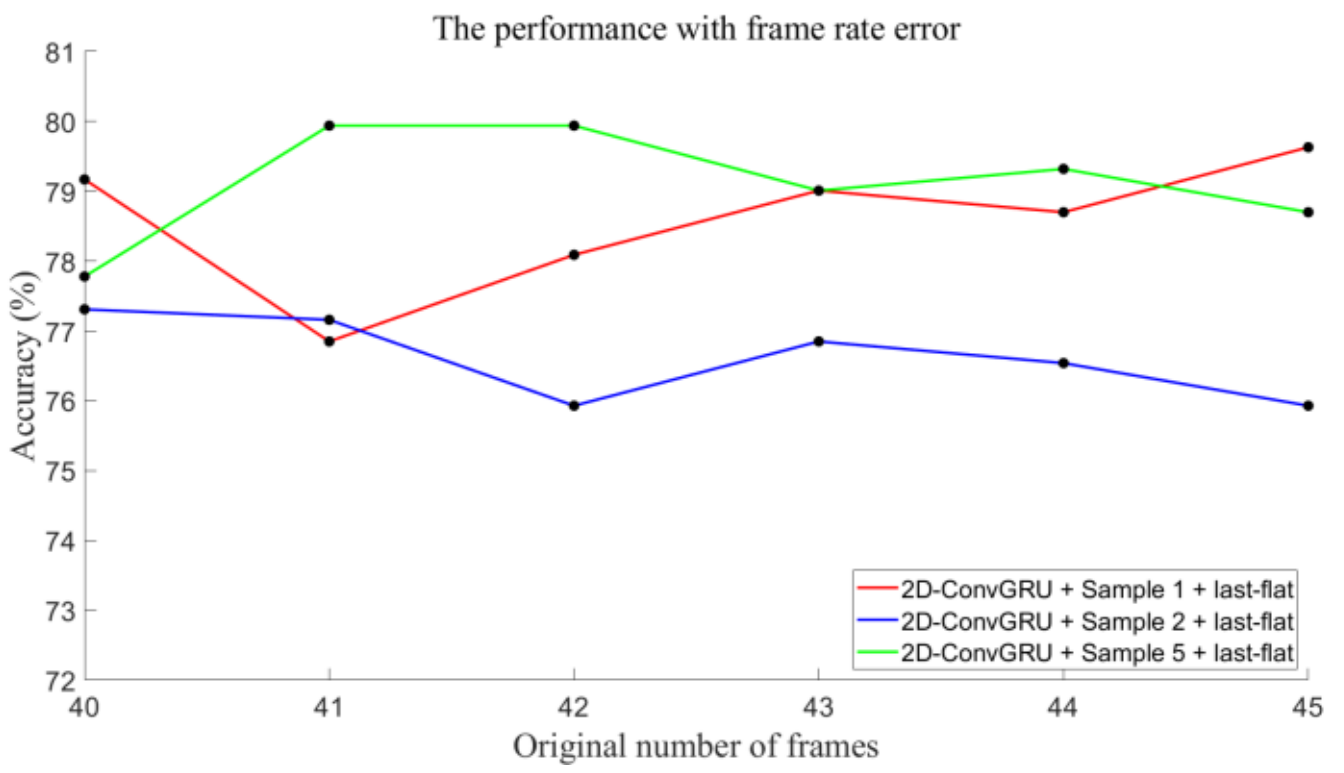

Fig.10. Accuracy of models with pre-trained models

In the second situation, we randomly set some images (1 to 5) in all 40 images as a $3 \times 224 \times 224$ zero tensor. In Fig.11, we can find out that all the models have robustness (over 75% accuracy) but the model with longer sequences and shorter time intervals is influenced more by the missing frames. The model with sampling method 5 is the best among the three models.

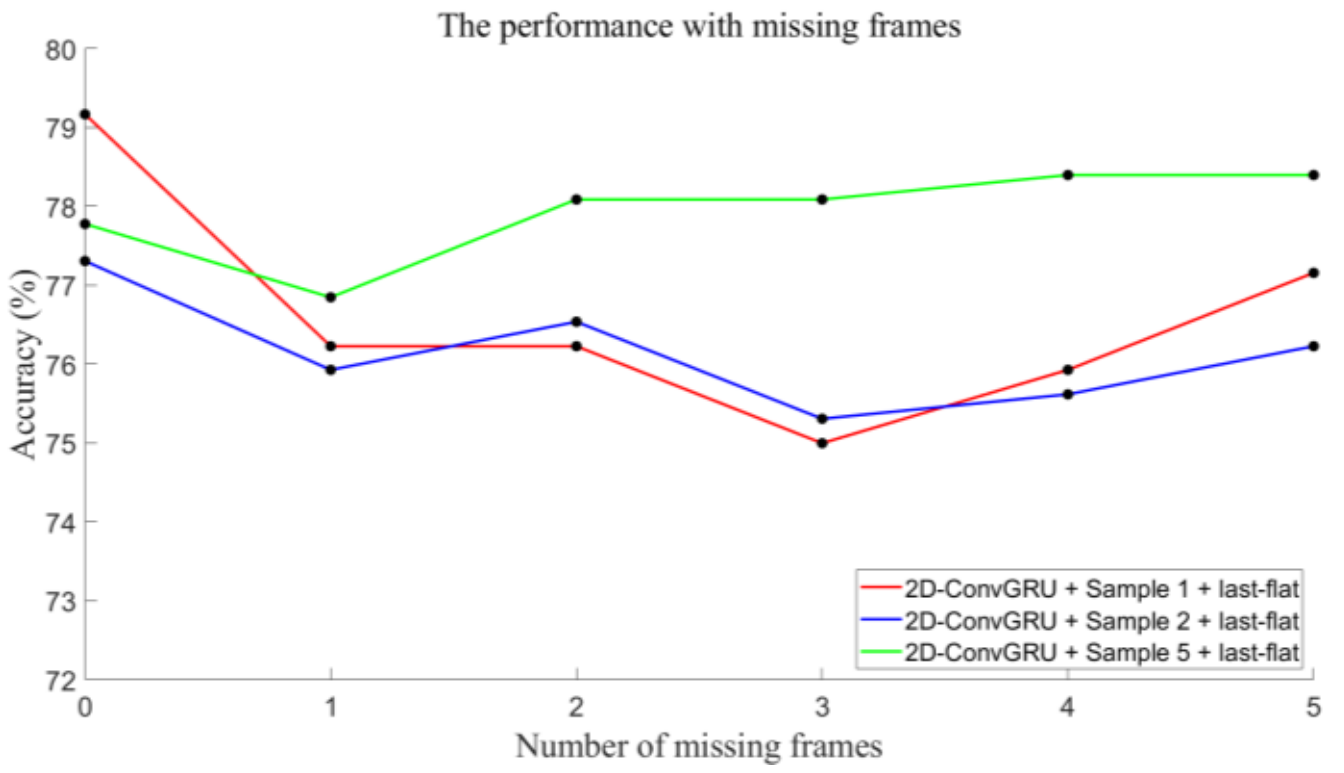

Fig.11. Accuracy of models with missing frames

In the last situation, the position errors, we randomly change the fall location to be off the center with a bias of 1 to 10 pixels. The results are shown in Fig.12. There is an apparent decrease of the accuracy in all three models, which indicates that the ConvGRU on spatial errors is not as robust as that with temporal errors. At the same time, the model with longer sequences and shorter time intervals is better than the other two. The reason is probably that more temporal information supplements the decrease spatial information caused by position errors.

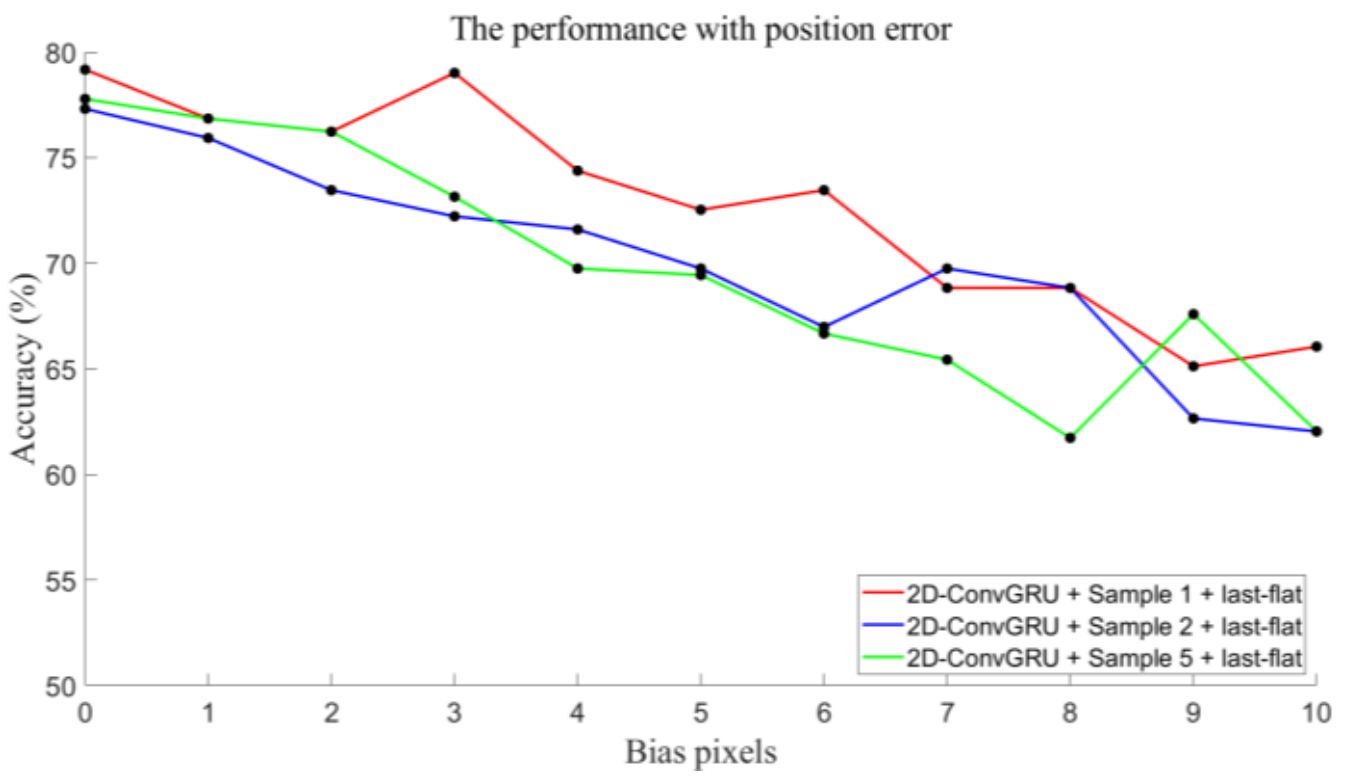

Fig.12. Accuracy of models with missing frames

## VI. CONCLUSION AND FUTURE WORK

In this paper, we achieved fine-grained action recognition using a dataset of human pitching a ball and an LSTM-based neural network architecture ConvGRU. A modified end-to-end AlexNet based model was incorporated and tested. Different experiment implementations were fully explored. Experiment results showed that our model achieved the state-of-the-art results, a maximum accuracy of 79.17%, which outperforms the previous two-stream methods and GRU based AlexNet. Our models also showed robustness against temporal (sampling rate and missing) and spatial (the location of the participant).

Although ConvGRU outperforms other models, we notice some limitations of ConvGRU. First, the sampling methods are limited. Since ConvGRU learns the temporal features from the image sequences, we need to keep the time interval a constant or near-constant and the time interval cannot be too large. Thus, the sampling methods are not flexible enough, which brings inconvenience to training. For example, TSN [30] can segment the video into 3 parts and randomly choose an image from each part for training, which does not work for ConvGRU. Second, it requires large RAM or VRAM. We assume a pitch trial with different length and test the memory occupation between a single $3 \times 3 \times 64$ convolutional layer in VGG16 and a 1D/2D-ConvGRU layer with the same size. The results show that when a new RGB image ($3 \times 224 \times 224$) is added to the sequence, a convolutional layer will occupy extra 24 MB memory, while 1D-ConvGRU 152 MB and 2D-ConvGRU 304 MB, which is more than 10 times of a convolutional layer. From Table 3 we can also see that the mean training time increases while an extra ConvGRU is added or the input sequence is lengthened. Considering the memory requirements, deep networks with multi-ConvGRU layer or input actions with long time is not recommended unless there are overwhelming computing resources.

In summary, ConvGRU is a high-accuracy and robust model for fine-grained action recognition, but it is only feasible for short duration actions and tiny networks.

As future work, we will address the following questions:

1) Is it worthy of using ConvGRU instead of deeper networks in fine-grained action recognition? If not, can the structure of ConvGRU be simpler?

2) How to apply ConvGRU to actions with different length? In this paper, our target actions are almost the same and so is the time duration of actions. However, some datasets require more flexibility, such as MPII-Cooking 2 [47], in which the time varies from 5 frames to several thousand frames.
3) How to better recognize the fine-grained actions in the full videos instead of a clip with a single action? We can consider some techniques about action segmentation in long-duration videos, such as those presented in [53].
4) Human actions are driven by intentions, however different from what they intend to do sometimes. How human intentions effect their actions and the outcomes and predict human intentions from the actions? Correctly predicting human intentions may further help robots to understand and work with people without being misled by some error actions.